\pdfoutput=1

\documentclass[11pt]{article}

\usepackage{ACL2023}

\usepackage{times}
\usepackage{latexsym}

\usepackage[T1]{fontenc}

\usepackage[utf8]{inputenc}

\usepackage{microtype}

\usepackage{inconsolata}
\usepackage{url}
\usepackage[linesnumbered,ruled,vlined]{algorithm2e}
\usepackage{multirow}
\usepackage{booktabs}
\usepackage{caption}
\usepackage{subcaption}
\usepackage{xspace}
\usepackage{amsmath}
\usepackage{amssymb}
\usepackage{newtxmath}
\usepackage{bbm}
\usepackage{graphicx}
\usepackage{tabularx}
\usepackage{bbding}
\usepackage{pifont}
\usepackage{diagbox}
\usepackage{makecell}
\usepackage{colortbl}
\usepackage{bbm}
\usepackage{color}
\usepackage{hhline}
\usepackage{soul}
\usepackage{float}

\newcommand{\paratitle}[1]{\vspace{1.5ex}\noindent\textbf{#1}}
\newcommand{\ie}{\emph{i.e.,}\xspace}

\newcommand{\eg}{\emph{e.g.,}\xspace}

\newcommand{\ignore}[1]{}

\title{Extracting and Combining Abilities For Building Multi-lingual Ability-enhanced Large Language Models}

\author{\textbf{Zhipeng Chen\textsuperscript{{1}},~
        Kun Zhou\textsuperscript{{2}},~
        Liang Song\textsuperscript{{3}},~
        Wayne Xin Zhao\textsuperscript{{1}}\thanks{\llap{}\:\:\:Corresponding authors.},~} \\
    \textbf{Bingning Wang\textsuperscript{{3}}\footnotemark[1],~
        Weipeng Chen\textsuperscript{{3}},~
        Ji-Rong Wen\textsuperscript{{1},{2}}
    } \\
    \textsuperscript{1}Gaoling School of Artificial Intelligence, Renmin University of China.\\
    \textsuperscript{2}School of Information, Renmin University of China.~~
    \textsuperscript{3}Baichuan Inc. \\
    \texttt{zhipeng\_chen@ruc.edu.cn,francis\_kun\_zhou@163.com}\\
    \texttt{batmanfly@gmail.com,daniel@baichuan-inc.com}
}

\begin{document}
\maketitle
\begin{abstract}
Multi-lingual ability transfer has become increasingly important for the broad application of large language models~(LLMs). 
Existing work highly relies on training with the multi-lingual ability-related data, which may not be available for low-resource languages. To solve it,  we propose a \textbf{M}ulti-lingual \textbf{A}bilities \textbf{E}xtraction and \textbf{C}ombination approach (\textbf{MAEC}), which decomposes and extracts language-agnostic ability-related weights from LLMs, and combines them across different languages by simple addition and subtraction operations without training. 
Specifically, our MAEC consists of the extraction and combination stages. In the extraction stage, we firstly locate \emph{key neurons} that are highly related to specific abilities, and then employ them to extract the transferable \emph{ability-related weights}. In the combination stage, we further select the \emph{ability-related tensors} that mitigate the linguistic effects, and design a combining strategy based on them and the \emph{language-specific weights}, to build the multi-lingual ability-enhanced LLM.
To assess the effectiveness of our approach, we conduct extensive experiments on LLaMA-3 8B on mathematical and scientific tasks in both high-resource and low-resource lingual scenarios. 
Empirical results have shown that MAEC can effectively and efficiently extract and combine the advanced abilities, achieving \textbf{comparable performance with PaLM}.
Resources are available at \url{https://github.com/RUCAIBox/MAET}.
\end{abstract}

\section{Introduction}

\begin{figure}[t]
    \centering
    \includegraphics[width=1\linewidth]{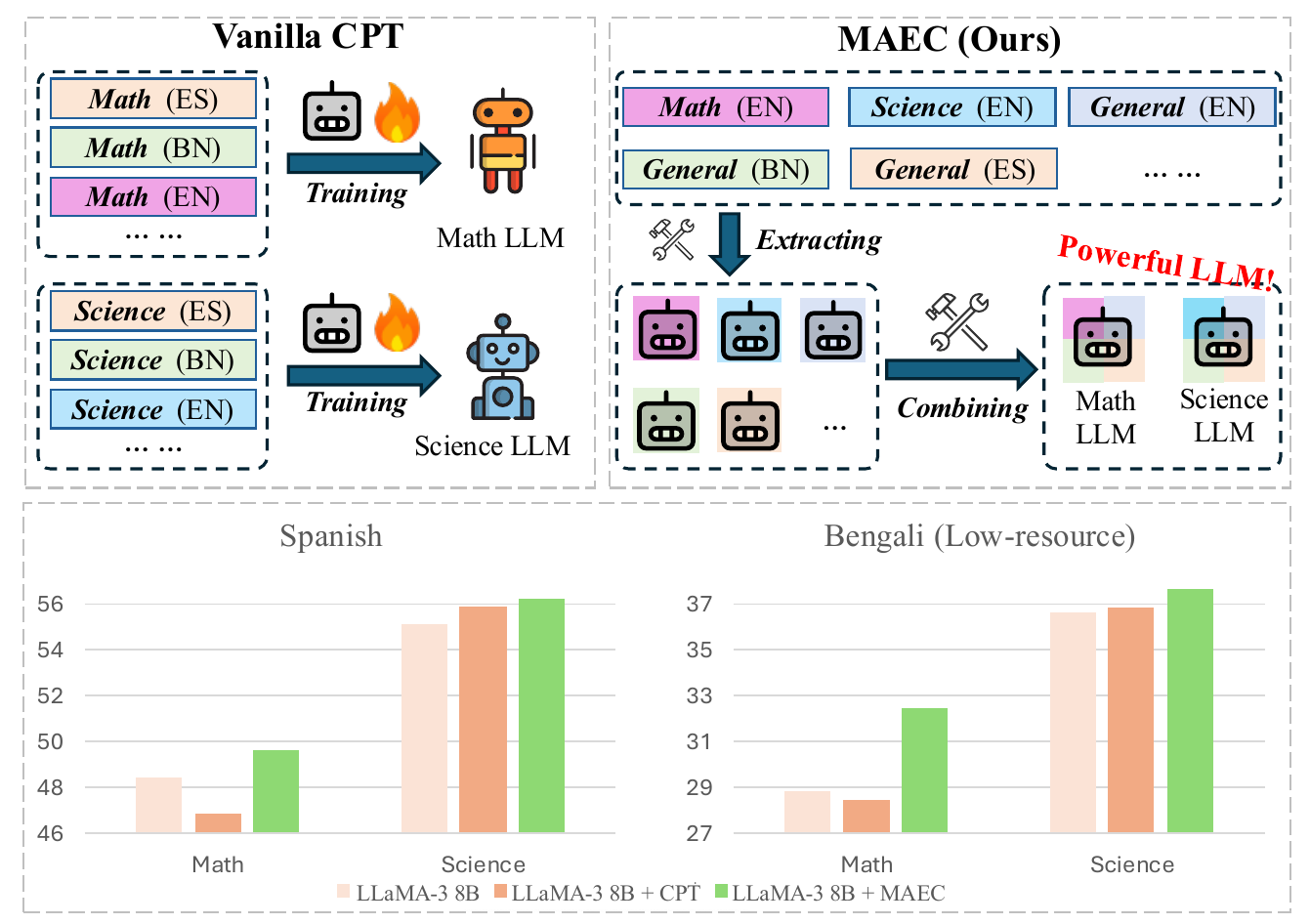}
    \caption{The comparison between CPT and MAEC. Only with the single-lingual ability-related corpus, MAEC can extract the abilities and combine them, achieving effective and efficient domain adaptation.}
    \label{fig-intro}
\end{figure}

Large language models~(LLMs) have shown remarkable performance on various general tasks, \eg text generation and question answering~\cite{llmsurvey,gpt4}.
Despite the success, LLMs are still struggling to solve complex tasks (\eg mathematical reasoning), which require LLMs to possess specific advanced abilities (\eg deductive reasoning) and knowledge (\eg mathematical theory)~\cite{mammoth2,scienceqa}.
To address it and further improve LLMs, existing work either collects the related data to train LLMs~\cite{Du-arxiv-2024-Unlocking,llama3_syne}, or merges the parameters of existing well-performed LLMs to transfer their abilities into one single model~\cite{task_vector,ties_merging}.

Despite the success, it is not easy to collect sufficient training corpus or well-trained LLMs related to specific abilities, especially in multi-lingual scenarios. Especially, some popular languages (\eg English) have dominated the linguistic expressions of the open web data, and the amount of available {domain-specific} data for low-resource languages (\eg Bengali or Telugu) is highly limited~{\cite{patzelt-arxiv-2024-medical,mirashi-arxiv-2024-importance}}. Fortunately, existing work~\cite{Zhao-arxiv-2024-LLaMA,Schafer-arxiv-2024-Language} has revealed that the learned knowledge from one language by LLMs could be inherited and leveraged by other languages.
For example, Llama-series LLMs are trained mainly on English texts, while they can also solve the tasks based on other languages. Such a finding has been widely explored in either improving the overall performance of multi-lingual LLMs~\cite{Schafer-arxiv-2024-Language} or enhancing fine-grained knowledge~\cite{llama3_syne}. 
However, the related work mostly requires the ability-related corpus in the target language, which is not always available for low-resource languages.

To conduct a more effective ability transfer, our idea is to learn and extract the ``\emph{ability-related weights}'' that preserves the knowledge about specific abilities for the LLM. 
If such ability-related and language-related weights could be decomposed, it is achievable to transfer the required abilities into target languages by just combining the corresponding weights, even building a multi-lingual ability-enhanced LLM like building blocks.
Based on this idea, in this paper, we propose a \textbf{M}ulti-lingual \textbf{A}bilities \textbf{E}xtraction and \textbf{C}ombination approach, named as \textbf{MAEC}.
Concretely, our approach consists of two major stages, \ie ability extracting and combining stage. 
In the extracting stage, we locate the abilities-related neurons and leverage the related corpus in a reference language to continually pre-train the LLM on the identified neurons. Then, based on the LLM trained on the general corpus, we devise a formula to extract the ability-related weights. 
In the combining stage, we utilize the ability-related weights to select related tensors, and design a specific merging strategy by interpolating linguistic and ability-related weights. 
As shown in Figure~\ref{fig-intro}, MAEC only needs ability-related corpus from any rich-resource language and multi-lingual general corpus, efficiently mitigating the data scarcity issues in low-resource languages. 

To assess the effectiveness of our approach, we conduct the evaluation based on two complex and comprehensive reasoning benchmarks, \ie Multi-lingual Grade School Math (MGSM)~\cite{mgsm} and science tasks from multi-lingual MMLU~\cite{okapi} as the evaluation benchmarks.
According to the evaluation results, with only training the specific LLM neurons on a small amount of data, the proposed approach MAEC outperforms other competitive baseline methods (\eg continual pre-training~\cite{cpt} and model merging methods with task vectors~\cite{task_vector}, achieving the 10\% relative improvement compared to the base LLM.

\section{Related Work}

\paratitle{Continual Pre-training.}
LLMs still struggle in complex tasks and low-resource lingual scenarios~\cite{Hedderich-NAACL-2021-ASurvey,deepseekmath}.
To adapt LLMs to a specific scenario, existing work~\cite{biogpt,galactica,SeaLLM3} has collected the related corpus to continually pre-train (CPT) LLMs.
During the CPT process, the mixture strategy between the general and ability-related corpus should be considered to avoid hurting their general abilities~\cite{data_mixing_laws,doremi,Siriwardhana-arxiv-2024-Domain}.
However, previous study~\cite {data_synthesis_survey,Lu-arxiv-2023-Machine} has found that it is hard to collect the task-related corpus, especially for low-resource language scenarios.
Therefore, synthesizing data from powerful LLMs is utilized to expand the task-related training corpus~\cite{chen-nature-2021-synthetic,jiuzhang3.0}.
In this work, we focus on adapting LLMs to multilingual complex reasoning scenarios with only the single-lingual ability-related corpus.

\paratitle{Knowledge Editing.}
According to the lottery ticket hypothesis~\cite{lottery_tickey}, training a sub-network of the model will achieve comparable or even better performance on downstream tasks.
Moreover, several study~\cite{allo,Zhang-ACL-2024-Unveiling} pointed out that the task-related sub-networks can be determined before the training process.
Existing study~\cite{Du-arxiv-2024-Unlocking,Wang-ACL-2024-Detoxifying,mixture_of_module} has leveraged the inner information of LLMs to select and train the related sub-network.
Besides, the probe (\ie a newly initialized parameter) can be implemented to detect the knowledge of LLMs and process targeted repair~\cite{Wang-arxiv-2024-Model,Jiang-ACL-2024-Learning}.

\paratitle{Model Merging.}
Given the huge computation resources consumed of CPT, previous work used model merging techniques to integrate different abilities (\eg mathematical reasoning and code synthesizing) into one model~\cite{Yang-arxiv-2024-Model,Xu-arxiv-2024-Training,Stoica-iclr-2024-ZipIt}.
During the merging process, the parameters of different LLMs might be conflict with others, which can be mitigated by the clip~\cite{ties_merging} or random dropout~\cite{dare} mechanism.
Moreover, the LLM inner parameters or external matrixes can be utilized to determine the hyper-parameters of the model merging process~\cite{Zhou-arxiv-2024-MetaGPT,Matena-neurips-2022-Merging}.
{Furthermore, existing work has merged the reasoning-specialized and multi-lingual models to improve their reasoning ability in non-English scenarios~\cite{MindMerger,LangBridge}.}
Inspired by the above work, we try to locate the task-related sub-networks of LLMs and transfer the advanced abilities.

\section{Preliminary}
Despite that LLMs exhibit remarkable performance on general tasks, they still have limited advanced abilities, \eg mathematical and scientific reasoning abilities. 
A typical approach to enhance these abilities is to continually pre-train~(CPT) LLMs with ability-related corpus. However, such training data might not always be available or sufficient, especially for minor domains (\eg Bengali). 
In this work, we focus on the task of \emph{ability extraction and transfer} by continual pre-training and merging LLMs. Concretely, LLMs are trained on the collected corpus from a certain domain, 
and we aim to only transfer its learned advanced capabilities to target domains~\citep{transfer_learning_survey_1,transfer_learning_survey_2} without further training.    
In this work, we study the cross-lingual scene where the linguistic-agnostic advanced ability and linguistic abilities should be extracted and transferred, to build a unified multi-lingual ability-enhanced LLM.

Formally, for a certain ability $A_i$ and a set of languages $L=\{L_0, L_1, \dots, L_n\}$, we assume that the general corpus of all languages can be collected, denoted as $\mathcal{C}_{\text{general}}=\{\mathcal{C}_{L_0}, \mathcal{C}_{L_1}, \dots, \mathcal{C}_{L_n}\}$, while the ability-related corpus is only available in language $L_0$ (\ie English), denoted as $\mathcal{C}_{L_0, A_i}$. 
Based on the above corpora, our goal is to extract and transfer the advanced ability $A_i$ from language $L_0$ and linguistic abilities from other languages $L_1, \dots, L_n$, into a unified LLM.

\section{Approach}

\begin{figure*}[t]
    \centering
    \includegraphics[width=1\linewidth]{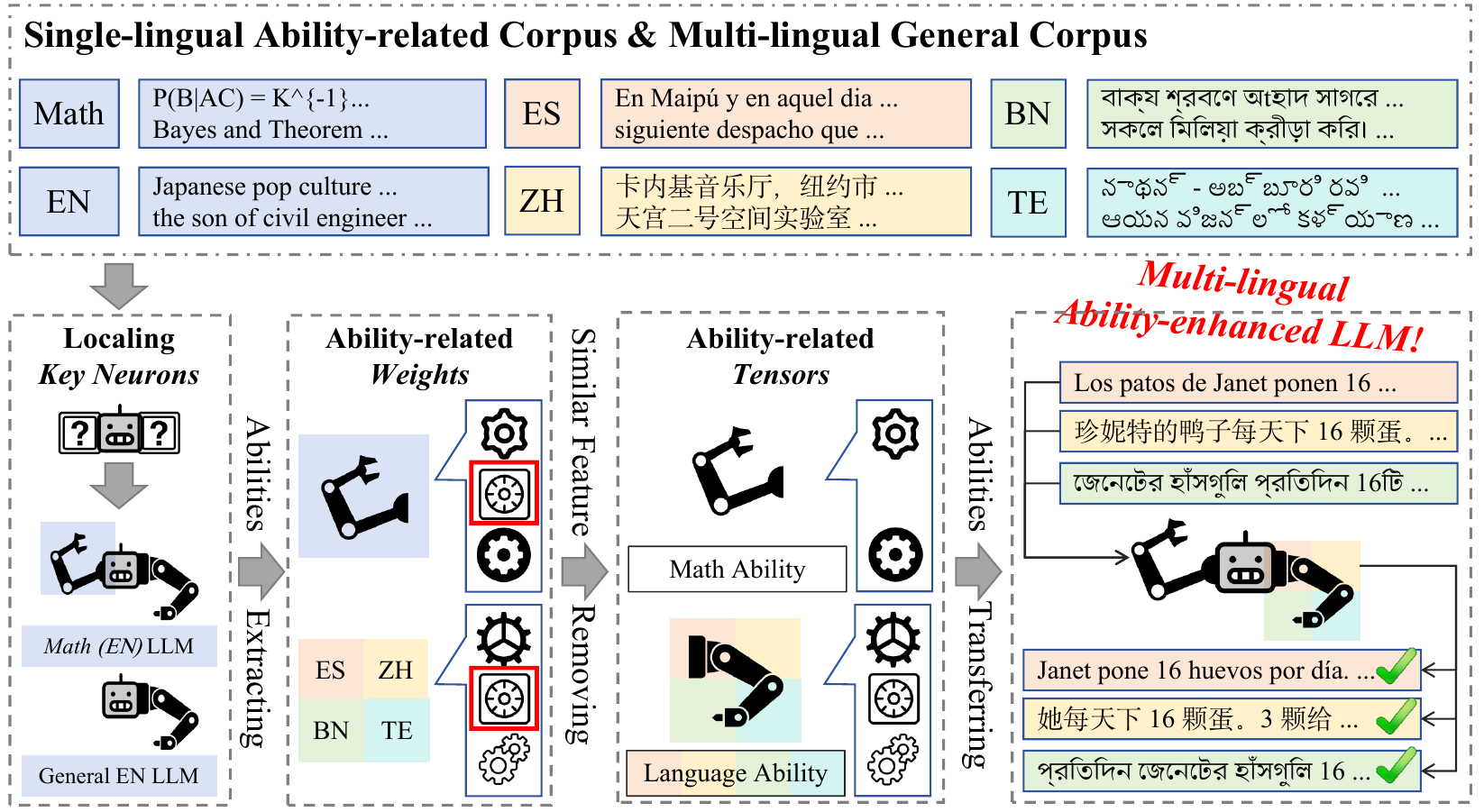}
    \caption{The framework of MAEC. First, we locate the key neurons, and utilize the single-lingual ability-related corpus and general corpus to train the LLM on these neurons to obtain the {ability-related} weight. Then, we remove the tensors related to language knowledge in the ability-related weight and combine the remaining to the base LLM. Finally, we obtain a powerful LLM that can solve the related tasks in multi-lingual scenarios.}
    \label{fig-framework}
\end{figure*}

In this section, we propose the \textbf{M}ulti-lingual \textbf{A}bility \textbf{E}xtraction and \textbf{C}ombination approach, named as \textbf{MAEC}, which can effectively transfer the advanced abilities from single-lingual LLMs, to build a multi-lingual ability-enhanced LLM. 
The key motivation of our approach is to identify and extract ability-related neurons or weights,
and combine the target abilities into a LLM in an efficient way.  
The framework of MAEC is presented in Figure~\ref{fig-framework}.

\subsection{Ability-related Weights Extraction}
\label{extraction}

In this part, we aim to locate and learn ability-related parameter weights within an LLM, to enable efficient combining of the ability into other LLMs.
Concretely, 
it consists of two major steps, \ie key neurons locating and ability-related parameter weights learning.  

\paratitle{Locating the Key Neurons.}
\label{local_key_neurons}
The gradient of each neuron in LLMs can be utilized to estimate its correlation degree with specific task ability~\cite{Pruthi-nips-2020-Estimating,allo,less}, we select those with high gradient values as key neurons.
To this end, we first use the ability-related corpus $\mathcal{C}_{L_0,A_i}'$, which denotes a subset of the original dataset $\mathcal{C}_{L_0,A_i}$, to continually pre-train the LLM, as sampling a small amount to train the model can be also applied to reduce the computation consumption.
During training, the LLM learns the language modeling task and each neuron is updated by the gradients associated by the training instances.
Due to the high cost of calculating the accumulation of gradient at each training step, we calculate the value changes of the LLM neurons before and after the training process to approximate the importance.
Formally, the importance function $I(A_i, \theta_j)$ of neurons can be computed as: 
\begin{equation}
\small
\label{eq-key_neuron}
    I(A_i, \theta_j) = \sum_{d_k \in \mathcal{C}_{L_0,A_i}'}{\text{Grad}\left(\theta_j, d_k\right)} \approx \frac{\parallel \tilde{\theta}_j-\theta_j \parallel}{\text{LearningRate}},
\end{equation}
where $d_k$ denotes the $k$-th instance of training corpus $\mathcal{C}_{L_0,A_i}'$ and $\tilde{\theta}_j$ denote the value of the $j$-th neuron of LLM after training, respectively.
{Based on it and inspired by previous work~\citep{ties_merging}, we rank all neurons according to their importance scores, and then select the top $k_1\%$ ones into the set $\mathcal{N}_{A_i}$ as the key neurons.}
Note that the trained model is only used for identifying the key neurons and is not used as the backbone for subsequent training processes.

\paratitle{Learning {Ability-related} Weights.}
Based on the identified key neurons in $\mathcal{N}_{A_i}$, we further learn the {ability-related} parameter weights.
Our motivation is to decompose the parameter weights according to their changes \emph{before} and \emph{after} the LLM has mastered a specific ability, which is achievable owing to the modularity and composition nature of the LLM parameter matrices~\citep{dare,mixture_of_expert}.
First, we utilize the key neurons locating method mentioned above to extract the ability-related neuron set $\mathcal{N}_{A_i}$, and also obtain the language-related neuron set $\mathcal{N}_{L_0}$ via the same way. 
Then, we train the LLM with the mixture of ability-related corpus and general corpus on the key neuron set $\mathcal{N}_{A_i}\bigcup\mathcal{N}_{L_0}$ and $\mathcal{N}_{L_0}$ respectively, to obtain two specific models, denoted as $\text{LLM}_{A_i,L_0}$ with parameters $\Theta_{A_i,L_0}$ and $\text{LLM}_{L_0}$ with parameters $\Theta_{L_0}$. 
Next, we measure the parameter changes between the backbone and the trained models, and obtain the {ability-related} weights via the parameter decomposition operation as:  
\begin{equation}
\small
\label{eq-extraction}
 R(A_i) = \alpha \cdot \underbrace{(\Theta_{A_i,L_0} - \Theta_o)}_{\text{Ability \& language difference}} - \beta \cdot \underbrace{(\Theta_{L_0} - \Theta_o)}_{\text{Language difference}},
\end{equation}
where $\alpha$ and $\beta$ are tunable coefficients to balance the two parts of weight differences, and $\Theta_o$ denote the original parameters of the LLM, which serves as the reference for parameter decomposition.  
As we only train the parameters within the neuron set, its weight difference should preserve the knowledge about the corresponding ability.
Thus, it can be regarded as the \emph{{ability-related} parameter representations}, and is promising to combine the ability into other LLMs by the addition operation.

\subsection{Multi-lingual Ability Combination}

After obtaining the ability-related weights, we combine them to transfer and integrate the abilities, building a multi-lingual ability-enhanced LLM.


\paratitle{Ability-related Tensor Selection.}
Although we can locate the ability-related key neurons, it is still hard to avoid the involvement of irrelevant ones.
Our empirical studies in Appendix~\ref{empirical_study} have found that neuron-level features are easy to be affected by the noisy data.
Inspired by previous work~\cite{cheng-etal-2024-enhancing}, we consider identifying ability-related tensors to further mitigate the linguistic effects, which correspond to the parameter matrices within the LLM.
Specifically, we firstly leverage the ability-related weights of languages $R(L_1), \dots, R(L_n)$ to obtain the multi-lingual weight $R_{Lang}$.
Given that large models have varying levels of proficiency in different languages, we use the hyper-parameters $\mu_1,\dots,\mu_n$ to tune this process as: 
\begin{equation}
\small
\label{eq-multilingual}
    R_{Lang} = \sum_{i=1}^{n}{\mu_i \cdot R(L_i)},
\end{equation}
where $R(L_i)$ preserves the linguistic ability of language $L_i$ learned based on Eq.~\ref{eq-extraction}. 
Therefore, $R_{Lang}$ can be considered as the general language ability of LLMs that spans multiple languages.
As we aim to find he parameter tensors that have low linguistic effects but focus on the desired abilities (\eg mathematical reasoning), we rank all the tensors according to their similarities with $R_{Lang}$, and pick up the last $k_2\%$ ones.
Formally, for tensor $\tau_i$,
we calculate the {cosine similarity} of this parameter between $R(A_i)$ and $R_{Lang}$, as follows,
\begin{equation}
\small
\label{eq-key_parameter}
    S(\tau_i) = \texttt{sim}\left(R(A_i)[\tau_i], R_{Lang}[\tau_i]\right),
\end{equation}
{
where we use the {cosine similarity} to implement the similarity function $\texttt{sim}\left(\cdot\right)$.
After obtaining the similarity of all tensors, we rank them in a descending order based on the similarity values, and then select the last $k_2\%$ parameters into the set $\mathcal{T}$ as the ability-related parameters.

\paratitle{Building Multi-lingual Ability-enhanced LLM.}
Based on the selected ability-related tensors $\mathcal{T}$, we design the model merging process by interpolating ability weights and multi-lingual weights, to build the multi-lingual ability-enhanced LLM. 
Formally, the final parameter tensors of the target LLM are computed as: 
\begin{equation}
\small
\label{eq-transfer}
\tilde{\tau}_i = \tau_i^{(o)} + \begin{cases} \gamma \cdot R(A_i)[\tau_i] + \eta \cdot R_{Lang}[\tau_i] ,&\tau_i \in \mathcal{T} \\ R_{Lang}[\tau_i],&\tau_i \notin \mathcal{T} \end{cases},
\end{equation}
where $\tau_i^{(o)}$ denotes the original value of parameter tensor $\tau_i$, and $\gamma$ and $\eta$ are tunable hyper-parameters.
This formula can be explained in two different cases. 
When a parameter tensor serves as the major role for specific abilities, we update it by adding both ability- and linguistic-related weights; otherwise, we 
simply enhance it with multi-lingual weights. 
In this way, we can derive a more powerful LLM that is equipped with the multi-lingual abilities and specific advanced abilities. 

\begin{table}[t]
    \small
    \centering
    \begin{tabular}{lcccc}
        \toprule
       \textbf{Approaches} & \textbf{MLAR} & \textbf{TPara} & \textbf{AC} & \textbf{AT} \\
        \midrule
        CPT & Yes & Full & No & No \\
        MoE & Yes & Full & No & No \\
        LoRA & Yes & Low-Rank & No & No \\
        MoL & Yes & Low-Rank & No & No \\
        TV & Yes & Full & Yes & No \\
        \midrule
        MAEC & \textbf{No} & \textbf{Ability-related} & \textbf{Yes} & \textbf{Yes} \\
        \bottomrule
    \end{tabular}
    \caption{The difference between our MAEC and the methods in previous work (\ie CPT~\citep{lora}, Mixture-of-Expert (MoE)~\citep{mixture_of_expert}, LoRA~\citep{lora}, Mixture-of-LoRA (MoL)~\citep{mixture_of_lora}, and Task Vector (TV)~\citep{task_vector}. MLAR, TPara, AC, and AT denote the abbreviation of multi-lingual ability-related corpus, parameters for training, ability composition, and ability transfer.}
    \label{tab-difference}
\end{table}
\subsection{The Overall Procedure}


To better demonstrate MAEC, we present {key concepts in Table~\ref{tab-concepts} for further clarifying} and provide the complete procedure in Algorithm~\ref{code_bub}. 
The procedure of MAEC consists of two main stages, \ie  ability-related weights extraction and multi-lingual ability combination.
For the extraction stage, we first utilize the accumulated gradient to estimate the importance of each neuron by Eq.~\ref{eq-key_neuron}. 
Then, we leverage the model trained on the general corpus to remove the effect of language and obtain the {ability-related} weight through Eq.~\ref{eq-extraction}.
In the combination stage, we utilize Eq.~\ref{eq-multilingual} and Eq.~\ref{eq-key_parameter} to obtain the multi-lingual weight and identify the ability-related tensors in LLM. 
After it, we leverage Eq.~\ref{eq-transfer} to fulfill the multi-lingual abilities combination, to build the multi-lingual ability-enhanced LLM. 

To highlight the difference between our approach and previous work, we present the comparison of these methods in Table~\ref{tab-difference}. 
To adapt LLMs to multi-lingual scenarios, most of the existing methods (\eg CPT and TV) require the multi-lingual ability-related corpus (\ie ability-related corpus is required for each language) for training the LLM. 
In comparison, our MAEC only trains and modifies the ability-related parameters, which can efficiently focus on enhancing the specific ability.
A major novelty of our work is that we identify the key units and implement the sparse update in the model training and merging procedure, which can effectively decompose, extract, and combine the abilities of LLMs. 
In addition, compared with the LoRA-based methods (\ie LoRA and MoL) that also sparsely update the LLM parameters, our approach selectively updates the ability-related neurons, while LoRA-based methods use the low-rank matrices to approximate the original parameters. 
\section{Experiment}

\begin{table*}[t]
    \small
    \centering
    \begin{tabular}{lccccc>{\columncolor{lightgray}}ccccc>{\columncolor{lightgray}}c}
        \toprule
        \multirow{2.5}*{\textbf{Methods}} & \multirow{2.5}*{\textbf{\#Tokens}} & \multicolumn{5}{c}{\textbf{Multilingual Mathematical Tasks}} & \multicolumn{5}{c}{\textbf{Multilingual Scientific Tasks}} \\
        \cmidrule(r){3-7}\cmidrule(r){8-12}
         & & ES & BN & TE & Avg. & ICER~($\downarrow$) & ES & BN & TE & Avg. & ICER~($\downarrow$)   \\
        \midrule
        \multicolumn{12}{c}{\textit{Close-source Multi-lingual Large Language Models}} \\
        GPT-3 175B & - & 54.8 & 10.8 & 4.8 & 23.5 & - & - & - & - & - & - \\
        PaLM 62B & - & 46.4 & 17.6 & 12.0 & 25.3 & - & - & - & - & - & - \\
        cont-PaLM 62B & - & 44.4 & 28.0 & 19.6 & 30.7 & - & - & - & - & - & - \\
        Flan-cont-PaLM 62B & - & 53.6 & 34.4 & 28.8 & 38.9 & - & - & - & - & - & - \\
        \midrule
        \multicolumn{12}{c}{\textit{Open-source Multi-lingual Large Language Models}} \\
        Baichuan-2 7B & - & 17.2 & 4.8 & 2.4 & 8.1 & - & 42.3 & 30.2 & 26.2 & 32.9 & -  \\
        Mistral 7B & - & 38.8 & 9.6 & 2.8 & 17.1 & - & 52.1 & 32.9 & 28.0 & 37.7 & -  \\
        LLaMA-2 7B & - & 7.6 & 1.6 & 0.0 & 3.1 & - & 34.2 & 24.6 & 22.2 & 27.0 & -  \\
        LLaMA-3 8B & - & 48.4 & 28.8 & 20.4 & 32.5 & - & 55.1 & 36.6 & 29.3 & 40.3 & -  \\
        \midrule
        \multicolumn{12}{c}{\textit{Vanilla Continually Pre-training based Approaches}} \\
        + $\text{F-CPT}_{\text{L\&A}}$ & 20B & 46.8 & 28.4 & 27.6 & \underline{34.3} & 11.1 & 55.9 & 36.8 & 30.1 & \underline{41.0} & 28.6 \\
        + $\text{L-CPT}_{\text{L\&A}}$ & 20B & 44.8 & 28.8 & 23.6 & 32.4 & - & 54.8 & 36.4 & 29.9 & 40.4 & 200.0 \\
        + {$\text{F-CPT}_{\text{A}}$} & {4B} & 47.2 & 20.0 & 13.2 & 26.8 & - & 51.9 & 33.4 & 29.4 & 38.2 & - \\
        + $\text{F-CPT}_{\text{L}}$ & 8B & 38.8 & 28.0 & 23.6 & 30.1 & - & 53.6 & 35.9 & 30.6 & 40.0 & - \\
        + $\text{L-CPT}_{\text{L}}$ & 8B & 46.4 & 28.4 & 22.8 & 32.5 & - & 55.0 & 36.7 & 30.4 & 40.7 & \underline{20.0} \\
        \midrule
        \multicolumn{12}{c}{\textit{Transfer Learning based Approaches}} \\
        + $\text{F-CPT}_{\text{L}}$ \& DA & 12B & 41.6 & 30.4 & 27.6 & 33.2 & 17.1 & 52.7 & 35.5 & 28.6 & 38.9 & - \\
        + $\text{L-CPT}_{\text{L}}$ \& DA & 12B & 46.8 & 28.0 & 27.2 & 34.0 & \underline{8.0} & 55.7 & 36.5 & 29.7 & 40.6 & 40.0 \\
        \midrule
        \multicolumn{12}{c}{\textit{Data Augmentation based Approaches}} \\
        + {$\text{F-CPT}_{\text{L\&T}}$} & 20B & 48.0 & 28.4 & 25.5 & 34.0 & 13.3 & 53.7 & 35.1 & 31.7 & 40.2 & - \\
        + {$\text{F-CPT}_{\text{T}}$} & 20B & 48.0 & 27.2 & 24.4 & 33.2 & 28.6 & 50.4 & 34.5 & 34.5 & 39.8 & - \\
        \midrule
        \multicolumn{12}{c}{\textit{Model Merging based Approaches}} \\
        + F-TV & {12B} & 42.0 & 16.0 & 10.4 & 22.8 & - & 53.4 & 36.7 & 30.7 & 40.3 & - \\
        + L-TV & {12B} & 45.6 & 30.8 & 25.6 & 34.0 & \underline{8.0} & 55.5 & 36.7 & 30.4 & 40.9 & \underline{20.0} \\
        \midrule
        + MAEC (Ours) & {12B} & 49.6 & 32.4 & 25.2 & \textbf{35.7} & \textbf{3.6} & 56.2 & 37.6 & 30.4 & \textbf{41.4} & \textbf{10.9} \\
        \bottomrule
    \end{tabular}
    \caption{The performance of different approaches on multilingual mathematical and scientific tasks. ES, BN, and TE denote Spanish, Bengali, and Telugu, respectively. \#Tokens denotes the number of training tokens. 
    }
    \label{main_results}
\end{table*}

\subsection{Experimental Settings}

We introduce the datasets, metrics, and the baselines in our evaluation, and present the implementation details of our approach in Appendix~\ref{sec-hyper_parameter}.

\paratitle{Datasets.}
We focus on transferring the advanced abilities (\ie mathematical and scientific reasoning abilities) of LLMs from English scenarios to multi-lingual scenarios, including high-resource language (\ie \emph{Spanish}) and low-resource languages (\ie \emph{Bengali} and \emph{Telugu}).
Thus, for the training corpus, we extract the corpus proposed by previous work~\citep{baichuan2,bloom} as the general corpus, and use \emph{OpenWebMath}~\citep{openwebmath} and \emph{arXiv papers}~\citep{dolma} as the ability-related corpus for mathematical and scientific tasks.
For evaluation, we follow the settings in previous work~\citep{gpt4}, utilizing {\emph{Multi-lingual Grade School Math (MGSM)}}~\citep{mgsm} and science tasks from {\emph{multi-lingual MMLU}}~\citep{okapi} (\ie college and high school biology, chemistry, and physics) as the downstream tasks.
The statistical information of the datasets is shown in Table~\ref{tab-dataset}.

\paratitle{Evaluation Metrics}
We calculate the accuracy of the predicted answers from LLMs and focus on \emph{the average performance (Avg.)}, since our major goal is building a multi-lingual LLM.
Moreover, we introduce \emph{the incremental cost-effectiveness ratio (ICER)}~\cite{ICER} to assess the efficiency of the approaches, \ie \texttt{ICER} = \#\texttt{Tokens} / \texttt{Improvement} $\times 100\%$.
Notably, we only report the ICER scores for the methods that can lead to improvements.

\paratitle{Baselines.}
We adopt \emph{LLaMA-3 8B}~\citep{llama3} as the backbone model and four categories of widely used methods as baselines, \ie \emph{continually pre-training}, \emph{transfer learning}, \emph{data augmentation}, and \emph{model merging based} approaches. 
Concretely, a baseline can be represented as three parts, \ie training parameters, training approach, and training data.
First, we conduct the full parameters training and the LoRA training~\citep{lora}, denoted as the \emph{``F''} and \emph{``L''} at the prefix, respectively.
Second, for the training approach, we employ {continual pre-training \emph{(CPT)}}~\citep{cpt}, {domain adaption \emph{(DA)}}~\citep{galactica}, and {model merging with task vector \emph{(TV)}}~\citep{task_vector}.
Third, for the training data, {\emph{``L'',  ``A'', and ``T''}} refer to the multi-lingual general corpus, English ability-related corpus, and multi-lingual ability-related corpus translated by GPT-4o~\citep{gpt4o}, respectively.
Also, we present the performance of open-source LLMs (\ie Baichuan-2 7B~\citep{baichuan2}, Mistral 7B~\citep{mistral}, and LLaMA-2 7B~\citep{llama2}) and close-source LLMs (\ie GPT-3 and PaLM series model~\citep{flan})

\subsection{Main Results}
The evaluation results have been shown in Table~\ref{main_results}.

First, MAEC outperforms other baselines in the average performance of all downstream tasks by only expensing 60\% computational resources, showing the best incremental cost effectiveness ratio.
In our experiment, continually pre-training LLMs on a mixture of multi-lingual general corpus and single-lingual ability-related corpus (\ie F-CPT$_{\text{L}\&\text{A}}$) can enhance the specific ability of LLMs, achieving the second-best performance.
However, when adapting LLMs to a new domain or enhancing a new ability of LLM, CPT-based methods should retrain the LLMs on the ability-related and multi-lingual corpus, showing the lack of transferability and requirements of more computational resources.
For the new domain adapting, MAEC only utilizes a small amount of single-lingual ability-related corpus (\ie English corpus in practice) to obtain the ability weight, which can be employed to combine the corresponding advanced ability, achieving both effectiveness and efficiency.


Second, although our MAEC shows similar training efficiency to transfer learning based approaches, MAEC performs better than these baselines, showing the lower ICER score (\eg 3.6 \emph{v.s.} 8.0).
For transfer learning based approaches, since the model is only trained on the single-lingual ability-related corpus during the domain adaptation process, it is difficult for LLM to handle the challenging tasks in multi-lingual scenarios.
Concretely, the performance of LLM on the multi-lingual scientific tasks even decreases after domain adaptation, showing a 4\% relative decrease.
To alleviate this issue, MAEC leverages the calculation between the parameters of different models to extract the ability-related weights, which are language-agnostic and can be transferred to any other scenario.

Third, MAEC also achieves higher performance than data augmentation based approaches (\ie training LLM on the multi-lingual ability-related corpus translated by GPT-4o).
The translation-based method consumes more computational resources and cannot achieve better performance. The reason might be that LLMs cannot perform the translation process well and the translated corpus shares similar knowledge of the specific domain, which makes LLM overfit the corresponding knowledge and cannot really understand the specific knowledge. In contrast, our approach decomposes the advanced ability and language ability, and transfers the advanced ability from one language to another, preventing overfitting, decreasing the expense, and improving performance.
These results demonstrate that data-centric methods are difficult to build a multi-lingual ability-enhanced LLM.

Last, compared with the model merging based approaches (\ie F-TV and L-TV), experimental results have shown that MAEC performs better than these baseline methods, since we decompose the relation between ability and the language of the training corpus.
In the previous model merging approaches, they mainly added the parameters of different models to obtain the final model, without considering the relation between language and abilities.
Due to the extraction mechanism of MAEC, we mitigate the effect of languages and make the weight more related to ability, which can be transferred in multi-lingual scenarios.

\subsection{Detailed Analysis}

To further analyze MAEC, we conduct an ablation study, and the analysis of the combining ratio $k_2$ and the generalization of MAEC.

\begin{figure}[t]
    \centering
    \includegraphics[width=1\linewidth]{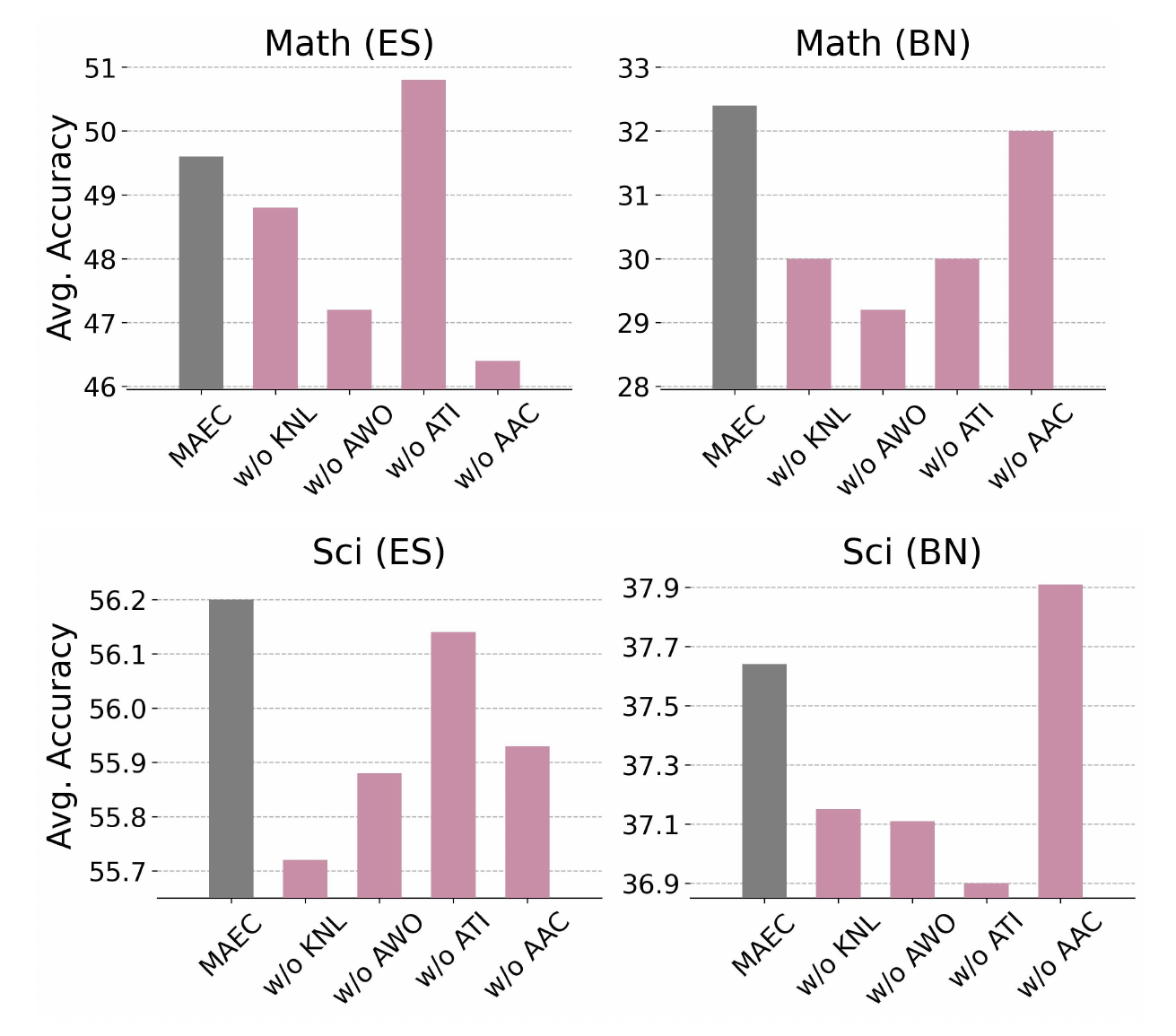}
    \caption{The ablation study. KNL, AWO, ATI, and AAC denote key neurons locating (Eq.~\ref{eq-key_neuron}), ability weights obtaining (Eq.~\ref{eq-extraction}), ability-related tensors identifying (Eq.~\ref{eq-key_parameter}), and advanced abilities combining (Eq.~\ref{eq-transfer}).}
    \label{fig-ablation}
\end{figure}

\paratitle{Ablation Study.}
To assess the effectiveness of each component of MAEC, we conduct the ablation study and present the results in Figure~\ref{fig-ablation}.
We implement MAEC on multi-lingual mathematical and scientific tasks without each module of MAEC, \ie key neurons locating (\ie Eq.~\ref{eq-key_neuron}), ability weight obtaining (\ie Eq.~\ref{eq-extraction}), ability-related parameter tensor identifying (\ie Eq.~\ref{eq-key_parameter}), and advanced abilities transferring (Eq.~\ref{eq-transfer}).
First, in most downstream scenarios, removing any module of MAEC will affect the final performance, verifying the effectiveness of the MAEC process.
Second, without ability weight obtaining, \ie directly utilizing the difference between LLM trained on the ability-related corpus and the backbone LLM as the ability weight, the performance of LLMs is seriously hurt in both scenarios, indicating this process can significantly extract the advanced abilities from the single-lingual corpus and decrease the influence of the language of the training corpus.
Third, comparing the results of the models whether adopting the ability transferring process, experimental results show that LLM with the multi-lingual ability-enhanced cannot well solve multi-lingual mathematical and scientific tasks, and leveraging the ability weight provided by MAEC can improve the LLM performance on advanced tasks.

\paratitle{Influence of Combining Ratio $k_2$.}
Identifying and updating the ability-related sub-network of LLMs is the key point of our MAEC.
We analyze the influence of the combining ratio $k_2\%$ and show the results in Figure~\ref{fig-ratio}.
Firstly, when the combining ratio $k_2$ changes within a certain range, the model's performance remains largely the same, indicating the strong robustness of our MAEC.
Specifically, for the mathematical tasks, when $k_2$ increases from 0.6 to 0.8, the performance of LLM remains approximately 35.5, showing the stability of MAEC.
Besides, the performance of LLM has decreased in both extremely low and high ratios of the ability-related parameters identifying process.
The main reason is that the lower proportion combines incomplete knowledge to the model and makes LLM unable to possess the corresponding ability, while the higher proportion cannot extract the ability weight precisely and will combine too much language-related knowledge to the model, which conflicts with the LLM's inner knowledge.

\begin{figure}[t]
    \centering
    \includegraphics[width=1\linewidth]{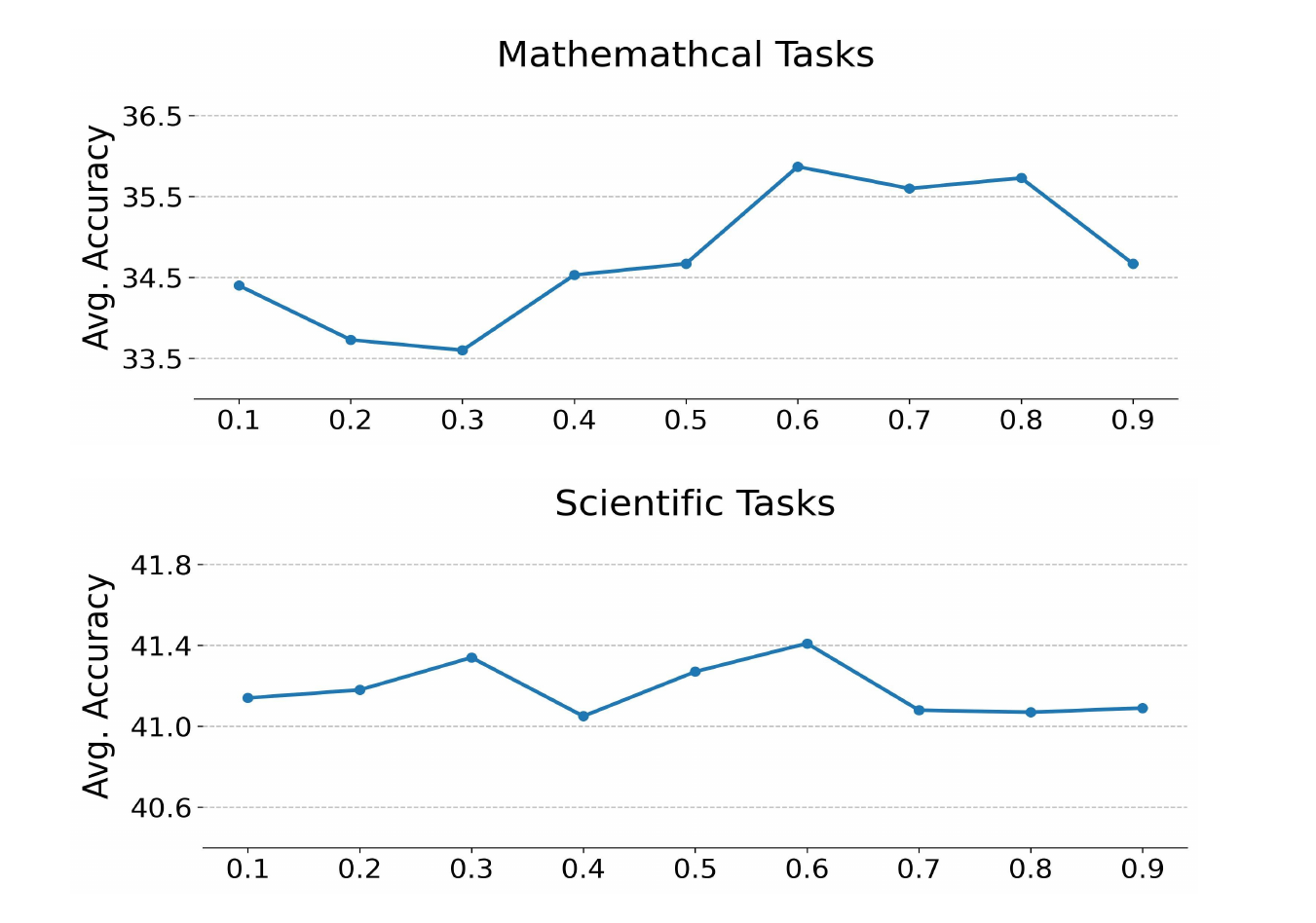}
    \caption{The performance of different proportions for the ability-related parameters identification.}
    \label{fig-ratio}
\end{figure}

\paratitle{Out-of-Domain Performance of MAEC.}
We conduct experiments about adapting mathematical ability on LLaMA-3 8B through MAEC, and assess its performance on out-of-domain (OOD) tasks (\ie MMLU~\citep{mmlu}, HumanEval~\citep{humaneval}, MBPP~\citep{mbpp}, and OpenbookQA~\citep{openbookqa}).
Results are presented in Table~\ref{generalization}.
We can observe that the performance of LLM on all evaluation tasks has decreased through the CPT training process, and the maximum decrease has been achieved 7.32\% on the HumanEval task.
One of the possible reasons is that LLaMA-3 has been trained on OpenWebMath during pre-training and the CPT process makes it overfit and forget the knowledge of other domains, hurting the performance on OOD tasks.
In contrast, our proposed MAEC achieves comparable and even better performance with backbone LLM in all downstream scenarios.
Since we identify and update the key neurons related to the specific ability, the ability of LLM can be precisely enhanced, and this strategy also helps the OOD tasks needed for mathematical ability, \eg mathematical tasks in MMLU and MBPP.

\begin{table}[t]
    \small
    \centering
    \begin{tabular}{lccccccc}
        \toprule
        \textbf{Methods} & \textbf{MMLU} & \textbf{MBPP} & \textbf{OpenbookQA} \\
        \midrule
        LLaMA-3 8B & 60.85 & 46.60 & 65.00 \\
        \midrule
        + CPT & \textcolor{red}{-2.39} & \textcolor{red}{-7.00} & \textcolor{red}{-3.60} \\
        + MAEC & +0.22 & {+0.80} & {+0.00} \\
        \bottomrule
    \end{tabular}
    \caption{The out-of-domain performance of different methods to train LLaMA-3 8B on OpenWebMath. After the ability-enhancing process, CPT hurts the OOD abilities of LLM, while MAEC can maintain these abilities.}
    \label{generalization}
\end{table}
\section{Conclusion}

In this paper, we presented MAET, which extracted the advanced ability-related weights from the LLM and supported simple addition and subtraction operations to transfer the ability across different languages.
Concretely, 
MAET included two main stages, \ie extraction and transfer.
For the extraction stage, we located the key neurons and extracted the ability-related weights.
Then, in the transfer stage, we identified the key parameter tensors and leveraged them to transfer the advanced ability into other LLMs.
In this process, the multi-lingual ability-related training corpus is not required, and the experimental results have shown that our approach outperformed competitive baselines.

As future work, we will consider better methods to identify the ability-related sub-network to decompose the abilities of LLMs and utilize an automated approach to determine the hyper-parameter.
Besides, we will implement MAET on larger-scale models, and scenarios with more languages and requiring more abilities to evaluate its effectiveness.

\section*{Limitations}
In this section, we discuss the limitations of our work.
First, we only implement our approach MAEC on 8B LLMs (\ie LLaMA-3 8B), and do not adopt the LLMs with larger scales (\eg 13B or 70B LLMs) in the experiment, due to the limitation of computational resources. We will test the effectiveness of our approach on these LLMs in the future.
Second, we only evaluate our approach on two downstream tasks (\ie mathematical and scientific reasoning tasks) in multi-lingual scenarios.
Although they are challenging and widely-used testbeds, it is still meaningful to verify our methods on other tasks.
Whereas, as we test the performance on diverse high-resource and low-resource languages, it can also provide comprehensive performance estimation for our approach in multi-lingual scenarios.
Finally, we do not consider the potential risk and ethics issues that might hurt the alignment of LLMs when using our approach.
Actually, our approach is also applicable to combining the ability to align across languages.
We will investigate to it in the future.

\section*{Acknowledgements}
This work was partially supported by National Natural Science Foundation of China under Grant No. 92470205 and 62222215, Beijing Natural Science Foundation under Grant No. L233008 and Beijing Municipal Science and Technology Project under Grant No. Z231100010323009. Xin Zhao and Bingning Wang are the corresponding authors.

\bibliography{anthology,custom}
\bibliographystyle{acl_natbib}

\newpage
\appendix


\begin{table*}[t]
    \small
    \centering
    \begin{tabular}{ll}
        \toprule
       \textbf{Concepts} & \textbf{Meaning} \\
        \midrule
        Key Neurons & \multicolumn{1}{l}{\begin{tabularx}{0.7\textwidth}{@{}X@{}}
            {Neuron refers to one of the trainable values of the tensors in LLMs. As previous work pointed out~\citep{Xu-arxiv-2024-Let}, different neurons might control the different abilities of LLMs. Following this finding, in our work, we define the neurons that control the specific ability as the ''Key Neurons''. Key neurons can be regarded as a set without duplication, and a neuron belonging to the set means that this neuron can control the specific ability~\citep{allo}. During the following training process, only the neurons belonging to the key neurons will be trained and optimized.}
        \end{tabularx}} \\
        \midrule
        Ability-related Weights & \multicolumn{1}{l}{\begin{tabularx}{0.7\textwidth}{@{}X@{}}
            {Ability-related weights refer to the value of the whole neuron in LLM, which can represent the corresponding ability of LLM~\citep{dare,task_vector}. In MAET, we obtain the ability-related weights through equation 2. The ability-related weights contain the value of all neurons. Since only the key neurons will be trained during the training process, the value of the neurons not belonging to key neurons is zero in the ability-related weights.}
        \end{tabularx}} \\
        \midrule
        Ability-related Tensors & \multicolumn{1}{l}{\begin{tabularx}{0.7\textwidth}{@{}X@{}}
            {Ability-related tensors can be regarded as a set of LLM tensors, which is related to the corresponding ability. Previous work has studied how the LLM layers influence the ability~\citep{Cheng-coling-2024-Enhancing}. Different from key neurons, ability-related tensors focus on higher-level information, integrating the sparse neurons into a coarser-grained element~\citep{xiao-arxiv-2024-configurable}. A tensor belonging to the ability-related tensors denotes that this tensor is highly related to the corresponding ability and can control this ability.}
        \end{tabularx}} \\
        \midrule
        Language-specific Weights & \multicolumn{1}{l}{\begin{tabularx}{0.7\textwidth}{@{}X@{}}
            {Similar to the ability-related weights, language-specific weights also refer to the value of the whole neurons in LLMs~\citep{Zhang-ACL-2024-Unveiling}. However, language-specific weights represent the language abilities of LLM that include multiple abilities (i.e., one language can be regarded as one ability)~\citep{Tang-ACL-2024-Language}, and the method of obtaining them is also different from ability-specific weights. In MAET, we first calculate the ability-related weights of each language and then Integrating these weights together to obtain the language-specific.}
        \end{tabularx}} \\
        \bottomrule
    \end{tabular}
    \caption{The key concepts of our approach.}
    \label{tab-concepts}
\end{table*}

\begin{algorithm*}[t]
\small
\caption{The complete procedure of our proposed approach MAET.}
\label{code_bub}
\SetKwInOut{Input}{Input}
\SetKwInOut{Output}{Output}

\Input{Single-lingual ability-related corpus $\mathcal{C}_{L_0,A_i}$, multi-lingual general corpus $\mathcal{C}_{L_0}, \mathcal{C}_{L_1},\dots, \mathcal{C}_{L_n}$, and the parameters of the backbone model $\Theta_o$.}
\Output{A well-trained multi-lingual ability-enhanced LLM.}

\BlankLine

\BlankLine
\tcp{Ability-related Weights Extraction}
$\theta' \leftarrow \text{CPT}(\mathcal{C}_{L_0,A_i}, \Theta_o$)\;

\For{$j$-th neuron in $\Theta_o$}{
    Calculate the importance score of the corresponding neuron using Eq.~\ref{eq-key_neuron}\;
}

Identify the key neuron set $\mathcal{N}_{A_i}$\;

$\text{LLM}_{A_i,L_0} \leftarrow \text{CPT}(\mathcal{C}_{L_0,A}, \Theta_o, \mathcal{N}_{A_i} \cup \mathcal{N}_{L_0})$\;

$\text{LLM}_{L_0} \leftarrow \text{CPT}(\mathcal{C}_{L_0}, \Theta_o, \mathcal{N}_{L_0})$\;

Learning the {ability-related} weight $R(A_i)$ using Eq.~\ref{eq-extraction}\;

\BlankLine
\tcp{Multi-lingual Ability Combination}

Obtaining the multi-lingual weight $R_{Lang}$ using Eq.~\ref{eq-multilingual}\;

\For{$j$-th parameter tensor in LLM} {
    Calculate the correlation using Eq.~\ref{eq-key_parameter}\;
}

Identify the ability-related parameters $\mathcal{T}$\;

Combine the ability to multi-lingual scenarios using Eq.~\ref{eq-transfer}\;

\BlankLine
Obtain the well-trained multi-lingual ability-enhanced LLM.

\end{algorithm*}

\begin{figure*}[t]
    \centering
    \subfloat[Loss During Training Process]{
        \includegraphics[width=0.31\linewidth]{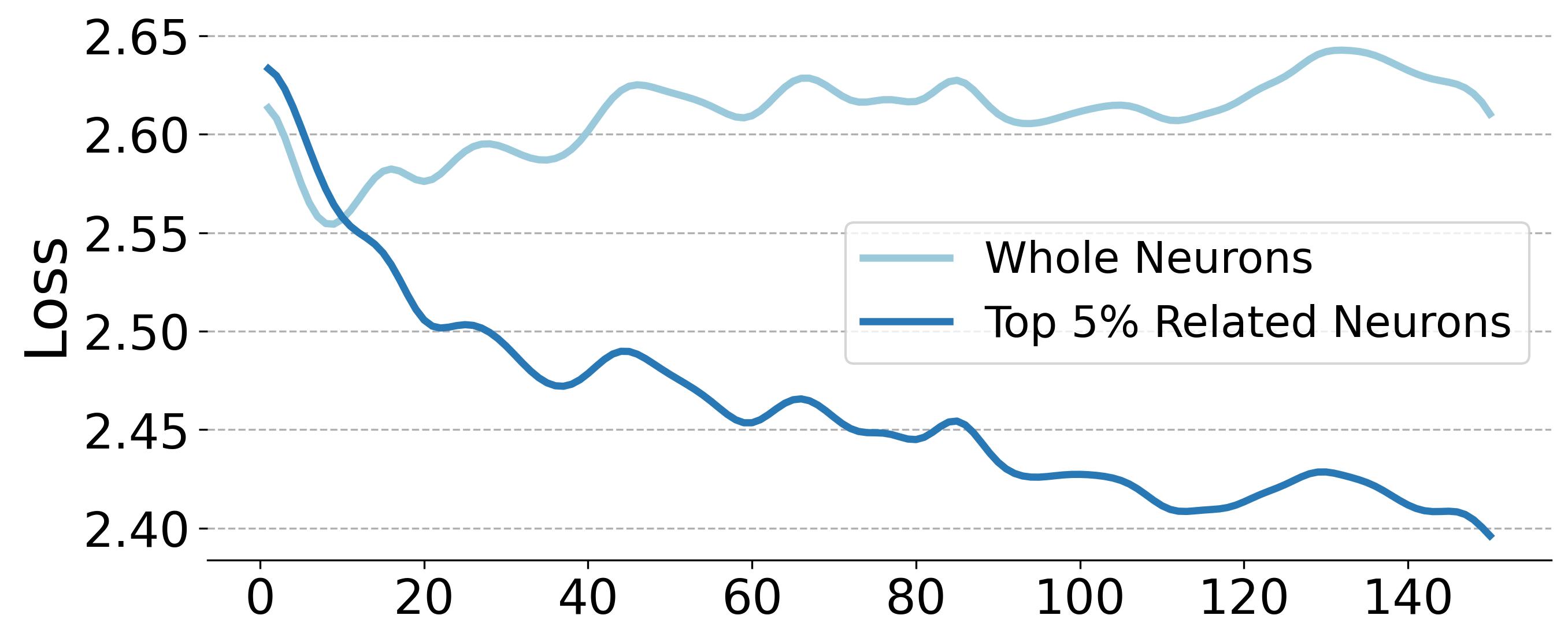}
        \label{empirical-loss}
    }
    \subfloat[Similarity of LLM Layers]{
        \includegraphics[width=0.31\linewidth]{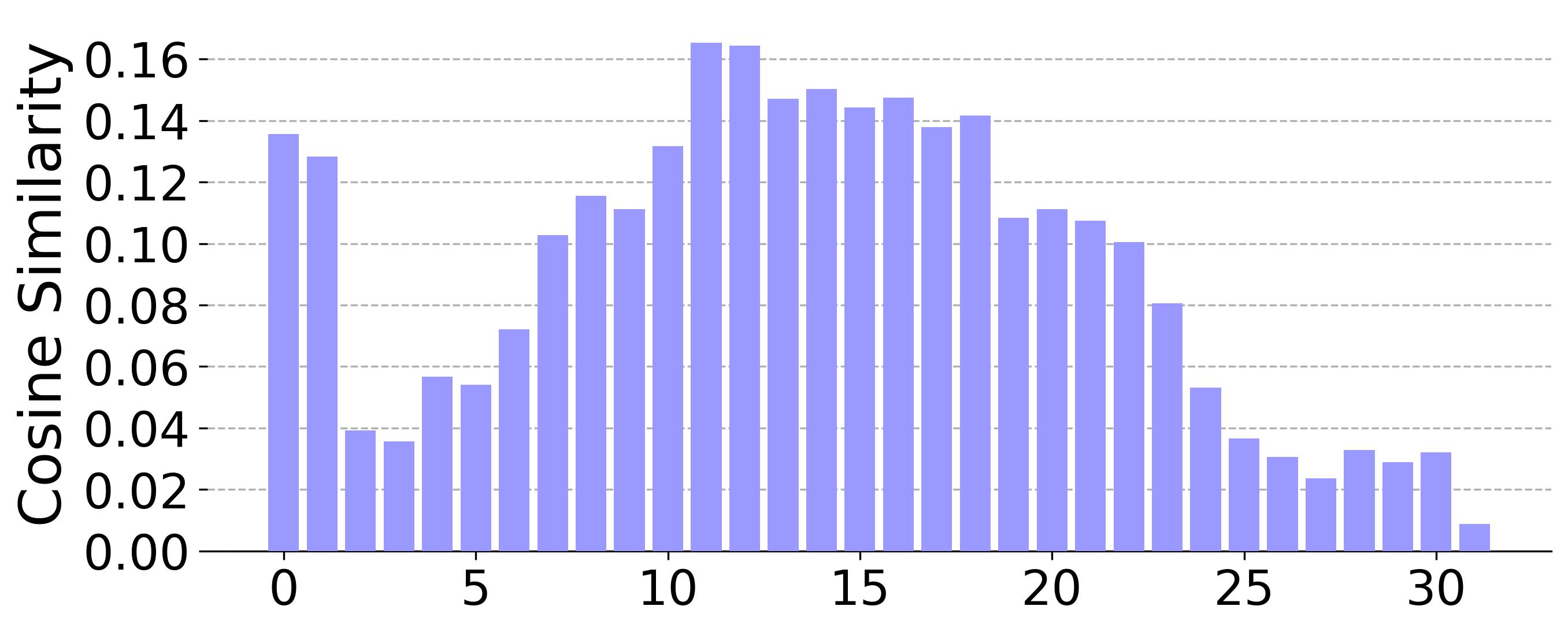}
        \label{empirical-layers}
    }
    \subfloat[Similarity of LLM Parameters]{
        \includegraphics[width=0.31\linewidth]{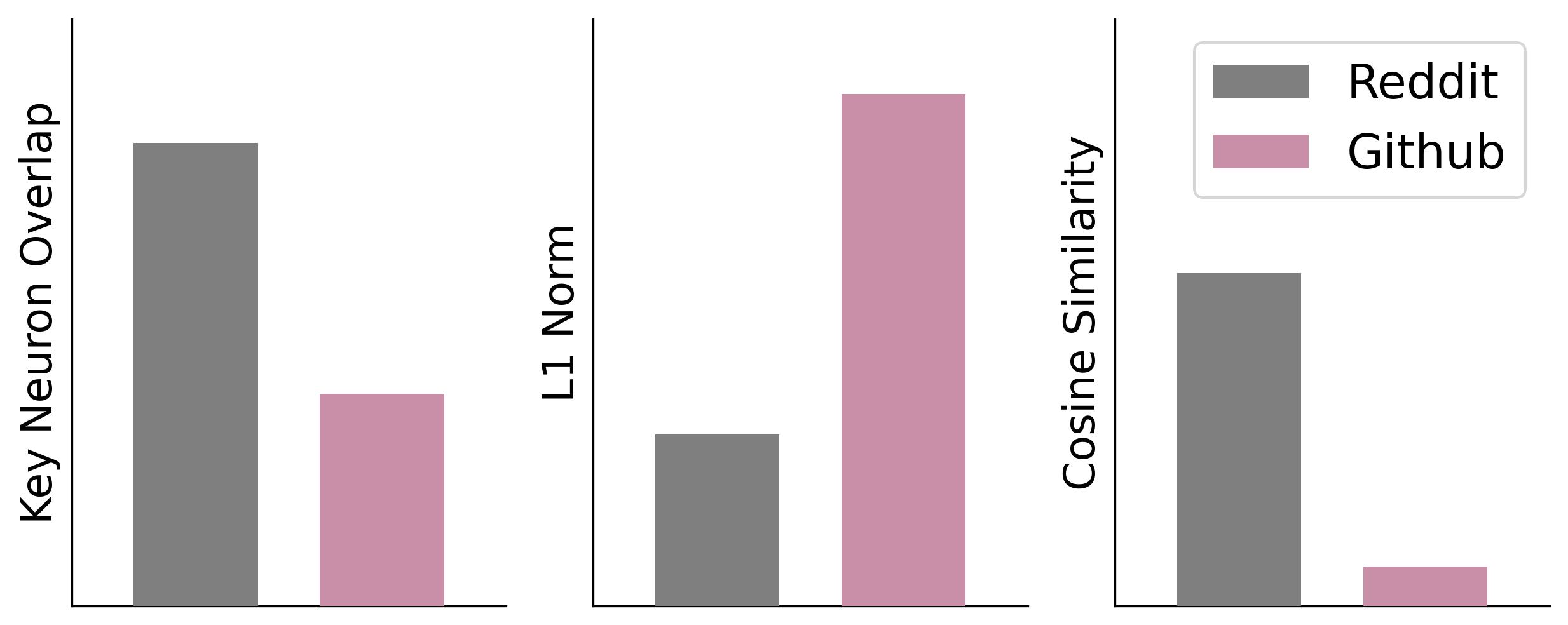}
        \label{empirical-similarity}
    }
    \caption{The results of empirical experiments. We present the loss of different training methods during the training process, the cosine similarity of LLM layers after being trained on Zhihu and Reddit, and the similarity of LLMs being trained on different training corpus.}
\label{empirical}
\end{figure*}

\section{Empirical Study}
\label{empirical_study}
A surge of work~\citep{Zhang-ACL-2024-Unveiling,xiao-arxiv-2024-configurable,Tang-ACL-2024-Language} has pointed out that LLMs sparsely activate the specific sub-modules to perform corresponding tasks.
Based on these findings, we conduct empirical experiments to explore whether the specific sub-module, which is related to advanced abilities, can be extracted and combined.
We utilize the forum corpus (\ie Zhihu for Chinese forum corpus and Reddit for English forum corpus) to continually pre-train LLMs, and then assess the training performance (\ie the value of loss function) and similarity of LLM neurons.

The forum corpus can be considered as containing the question-answering (QA) ability, which is necessary and important for LLMs.
The results from Figure~\ref{empirical-loss} have shown that only training the top 5\% relevant neurons of LLMs can achieve the lower training loss and fit into the training set more quickly, indicating that LLMs contain the sub-module corresponding to the QA ability.
Moreover, from Figure~\ref{empirical-layers} and Figure~\ref{empirical-similarity}, we can observe that the LLM trained on Zhihu has shown higher similarity with the LLM trained on Reddit than the LLM trained on Github (\ie lower L1 Norm and higher cosine similarity), and the cosine similarity of different layers in LLM are largely different.

According to the above results, we have found that the different sub-networks of LLMs control the different abilities, and precisely selecting the correct sub-module of LLMs will help the extraction of advanced abilities from the single-lingual corpus and the combination of these abilities to multi-lingual scenarios.
Concretely, although Zhihu and Reddit are in different languages, they will influence the similar sub-modules of LLM and make these sub-networks show high similarity with each other. These sub-networks can be referred to the ability-related sub-networks, which are slightly influenced by languages.

\begin{table*}[ht]
    \small
    \centering
    \begin{tabular}{cccc}
        \toprule
        Stage & Hyper-Parameter & Mathematical Tasks & Scientific Tasks \\
        \midrule
        \multirow{6}*{{Extraction}} & Learning Rate & $5\times 10^{-5}$ & $5\times 10^{-5}$  \\
        & Batch Size & 1M Tokens & 1M Tokens \\
        & Training Steps & 2B Tokens & 2B Tokens \\
        & $\alpha$ in Extraction & 0.8 & 0.8  \\
        & $\beta$ in Extraction & 0.2 & 0.2  \\
        & Ratio of Key Neurons $k_1$ & 5\% & 5\% \\
        \midrule
        \multirow{9}*{{Combination}} & Learning Rate & $5\times 10^{-5}$ & $5\times 10^{-5}$  \\
        & Batch Size & 1M Tokens & 1M Tokens \\
        & Training Steps & 2B Tokens & 2B Tokens \\
        & $\gamma$ in Combining & 0.2 & 0.2  \\
        & $\eta$ in Combining & 1.0 & 1.0  \\
        & Ratio of Key Tensors $k_2$ & 60\% & 60\% \\
        & {$\mu$ for Spanish} & {1.5} & {1.5} \\
        & {$\mu$ for Bengali} & {1.2} & {1.2} \\
        & {$\mu$ for Telugu} & {1.2} & {1.2} \\
        \bottomrule
    \end{tabular}
    \caption{The details of hyper-parameters in the training and evaluation process.}
    \label{hyper_parameters}
\end{table*}

\begin{table*}[ht]
\small
    \centering
    \begin{tabular}{lcccc}
        \toprule
        \multirow{2.5}*{\textbf{Language}} & \multicolumn{2}{c}{\textbf{Training Dataset (Tokens)}} & \multicolumn{2}{c}{\textbf{Evaluation Dataset (Instances)}} \\
        \cmidrule(r){2-3}\cmidrule(r){4-5}
        & General Corpus & Ability-related Corpus & Mathematical Tasks & Scientific Tasks \\
        \midrule
        English & 1.81B & 1.30B (Math) / 1.82B (Sci) & 250 & 1,245 \\
        Spanish & 1.81B & - & 250 & 1,232 \\
        Chinese & 1.80B & - & 250 & 1,229 \\
        Bengali & 1.81B & - & 250 & 1,137 \\
        Telugu & 1.81B & - & 250 & 1,036 \\
        \bottomrule
    \end{tabular}
    \caption{The statistical information of the training and evaluation datasets.}
    \label{tab-dataset}
\end{table*}

\begin{table*}[ht]
\small
    \centering
    \begin{tabular}{lcccccccccc}
        \toprule
        \multirow{2.5}*{\textbf{Methods}} & \multicolumn{3}{c}{\textbf{Qwen2.5 0.5B}} & \multicolumn{3}{c}{\textbf{Gemma2 2B}} \\
        \cmidrule(r){2-4}\cmidrule(r){5-7}
         & ES & TE & Avg. & ES & TE & Avg.   \\
        \midrule
        Backbone LLM & 36.64 & 25.69 & 31.17 & 43.41 & 30.01 & 36.71 \\
        \midrule
        + $\text{F-CPT}_{\text{L\&A}}$ & 32.90 & 22.43 & 27.67 & 38.48 & \textbf{30.39} & 34.62 \\
        + $\text{F-CPT}_{\text{A}}$ & 32.62 & 25.26 & 28.94 & 37.83 & 25.39 & 31.61 \\
        \midrule
        + MAET w/o API & 36.72 & 28.91 & 32.82 & 43.23 & 29.59 & 36.41 \\
        + MAET (Ours) & \textbf{36.91} & \textbf{29.62} & \textbf{33.27} & \textbf{43.62} & 30.37 & \textbf{37.00} \\
        \bottomrule
    \end{tabular}
    \caption{{The performance comparison of different LLMs on multilingual scientific tasks.}}
    \label{results_small_llm}
\end{table*}

\section{Implementation Details}
\label{sec-hyper_parameter}
In the experiment, we adapt LLaMA-3 8B as the backbone LLM, and employ \texttt{Transformers}~\citep{wolf-etal-2020-transformers} and \texttt{Deepspeed} framework to perform the training process.
{And we also present the evaluation results of different backbone LLM (\ie Qwen2.5 0.5B~\citep{qwen2.5} and Gemma2 2B~\citep{gemma2}) in Appendix~\ref{sec-small_llm}.}
For the training process, the learning rate, batch size, and training step are set as $5\times 10^{-5}$, $1$M tokens, and $2$B tokens, respectively.
Besides, for the key neurons locating, we select the top 5\% relevant neurons as the key neuron set $\mathcal{N}$ for both stages and identify the last 80\% and 60\% similar tensor as the key sub-network $\mathcal{T}$ for mathematical reasoning tasks and scientific reasoning tasks respectively.

\paratitle{Hyper-parameters Selection.}
we released all of the hyper-parameters during our experiment in Table~\ref{hyper_parameters}, to reproduce our proposed approach better. 
The hyperparameters discussed in the paper can be categorized into two types: training-related parameters (\eg learning rate, batch size) and training-independent parameters (\ie $\alpha$, $\beta$, $\gamma$, $\eta$, and $\mu$).
Training-related parameters do not require extensive hyperparameter tuning, as existing studies~\cite{llama3,qwen2.5} provide clear guidelines for setting them.
On the other hand, training-independent parameters are used to construct ability-related weights, tensors, and language-specific weights. These techniques are similar to those employed in model merging~\cite{task_vector,ties_merging}, and the hyperparameter setting approach outlined in the paper can be applied. A limited number of hyperparameter sets can be defined and validated, as the process primarily involves simple additions and subtractions of model parameters, making it computationally inexpensive.

\section{Details of Dataset}

We present the statistical information of the datasets in Table~\ref{tab-dataset}.
We mainly consider English, Spanish, Chinese, Bengali, and Telugu in our experiment, and utilized English as the in-domain language while others as the out-of-domain languages.
For the evaluation datasets, we select MGSM and multi-lingual MMLU as the evaluation benchmarks, which contain the parallel data in different languages and are useful for multi-lingual complex tasks evaluation.

{
\section{Prompt for Translation}
\label{translation_prompt}
\texttt{You should translate the following text from English to \{TARGET LANGUAGE\} and should not modify the latex code or website code. You should not add any details that are not mentioned in the original text.\\\\ \#\# English\\ \{ENGLISH TEXT\}\\\\ \#\# \{TARGET LANGUAGE\}}
}

{
\section{Performance of Small Scale LLMs}
\label{sec-small_llm}
We conduct the different LLMs with different sizes (\ie, Qwen2.5-0.5B and Gemma2-2B) in our experiment to valid the practicality of our approach. 
We assess MAET and baselines on multi-lingual scientific reasoning tasks and present the evaluation results in Table~\ref{results_small_llm}.
Comparing the performance of MAET and the baseline methods, we can observe that MAET can also enhance the performance of small scale models and outperform competitive baselines. Therefore, the evaluation results have shown the effectiveness of MAET and verified that MAET is a general LLM enhancement technology.
}

\begin{table*}[t]
\small
    \centering
    \begin{tabular}{lccccccccc}
        \toprule
        \textbf{LLM Tensors} & \textbf{Proportion of }$\mathcal{T}$ & \textbf{ES} & \textbf{ZH} & \textbf{BN} & \textbf{TE} & \textbf{Avg.} \\
        \midrule
        All Tensors & 100.00\% & 49.60 & 41.60 & 32.40 & 25.20 & 37.20 \\
        \midrule
        Attention All & 45.26\% & 48.80 & 41.60 & 28.80 & 26.40 & 36.40  \\
        \quad Attention Q & 12.07\% & 47.60 & 40.80 & 30.80 & 26.40 & 36.40  \\
        \quad Attention K & 10.34\% & 47.20 & 42.40 & 29.60 & 24.40 & 35.90  \\
        \quad Attention V & 9.48\% & 47.60 & 42.40 & 28.80 & 25.20 & 36.00  \\
        \quad Attention O & 13.36\% & 48.00 & 40.40 & 30.80 & 27.20 & 36.60  \\
        \midrule
        MLP All & 41.38\% & 48.80 & 39.60 & 31.60 & 27.60 & 36.90  \\
        \quad MLP Up & 13.79\% & 50.00 & 40.00 & 28.80 & 25.20 & 36.00  \\
        \quad MLP Gate & 13.79\% & 46.00 & 41.20 & 30.00 & 24.00 & 35.30  \\
        \quad MLP Down & 13.79\% & 49.60 & 41.60 & 30.40 & 26.00 & 36.90  \\
        \bottomrule
    \end{tabular}
    \caption{The effect of only merging the specific LLM tensors during the Combining process (\ie Eq.\ref{eq-transfer}) on multi-lingual mathematical tasks.}
    \label{sub_network}
\end{table*}

\section{Ability-related Sub-networks of LLM}
To assess and probe the ability-related sub-networks of LLMs, we only combine the specific tensors (\ie tensors in self-attention and MLP mechanism) from the ability weight to the final models through Eq.~\ref{eq-transfer}, to analyze the LLM inner abilities.
The experimental results are presented in Table~\ref{sub_network}.
From the experiment, we can observe that although the proportion of MLP layers (41.38\%) is lower than the attention layers (45.26\%), only combining the MLP layers outperforms Combining the attention layers, indicating that the MLP layers are more related to the advanced abilities and stores the corresponding knowledge.
In the MLP layers of LLM, the gate mechanism (\ie MLP Gate) will control the transmission of information and the down project mechanism (\ie MLP Down) will integrate the knowledge from previous layers, so that Combining the MLP layers can achieve better performance on the downstream tasks.



\section{Further Explanation of Baselines}

To better introduce the baseline methods, we propose a table to explain the details of each baseline in Table~\ref{explan_baseline}.

\begin{table*}[t]
\small
    \centering
    \begin{tabular}{lcccc}
        \toprule
        \textbf{Baselines} & \textbf{Training Parameters} & \textbf{Optimization Method} & \textbf{General Corpus} & \textbf{Ability-relared Corpus} \\
        \midrule
        F-CPT$_{L \& A}$ & Full Parameters & CPT & Multi-lingual & Single-lingual \\
        L-CPT$_{L \& A}$ & LoRA & CPT & Multi-lingual & Single-lingual \\
        F-CPT$_{A}$ & Full Parameters & CPT & No & Single-lingual \\
        F-CPT$_{L}$ & Full Parameters & CPT & Multi-lingual & No \\
        L-CPT$_{L}$ & LoRA & CPT & Multi-lingual & No \\
        F-CPT$_{L \& T}$ & Full Parameters & CPT & Multi-lingual & Translated Multi-lingual \\
        F-CPT$_{T}$ & Full Parameters & CPT & No & Translated Multi-lingual \\
        F-TV & Full Parameters & Model Merging & Multi-lingual & Single-lingual \\
        L-TV & LoRA & Model Merging & Multi-lingual & Single-lingual \\
        \bottomrule
    \end{tabular}
    \caption{The detailed explanation of each baseline mentioned in Table~\ref{main_results}.}
    \label{explan_baseline}
\end{table*}


\end{document}